\documentclass[letterpaper,journal]{IEEEtran}
\usepackage{amsmath,amsfonts}
\usepackage{algorithmic}
\usepackage{algorithm}
\usepackage{array}
\usepackage{tabularx}
\usepackage[caption=false,font=normalsize,labelfont=sf,textfont=sf]{subfig}
\usepackage{textcomp}
\usepackage{stfloats}
\usepackage{url}
\usepackage{verbatim}
\usepackage{graphicx}
\usepackage{cite}
\usepackage{setspace}
\usepackage[T1]{fontenc}
\usepackage[utf8]{inputenc}
\usepackage[normalem]{ulem}
\usepackage[hidelinks]{hyperref}

\usepackage{color}
\newcommand{\red}[1]{\textcolor{red}{#1}}

\newcommand{\edit}[2]{\red{\sout{#1}}\ifx\empty#1\empty\else\ifx\empty#2\empty\else\space\fi\fi\red{#2}}

\begin{document}
\bstctlcite{rollbot-reference-style}

\title{Rollbot 2: Agile Locomotion by a Single-Actuator Spherical Robot}

\author{Jingxian Wang*, Chen Yu*, Sam Kriegman, and Michael Rubenstein%
\thanks{Jingxian Wang and Chen Yu contributed equally to this work.}%
\thanks{Jingxian Wang, Chen Yu, Sam Kriegman, and Michael Rubenstein are with Northwestern University, Evanston, IL 60208 USA.}}

\markboth{IEEE Transactions on Robotics}{}

\maketitle

\begin{abstract}
Reducing actuator count simplifies robot hardware but also removes independent inputs for maneuvering. The original Rollbot steered by modulating a quasi-stable helical orbit, but moved slowly and relied on external motion capture. Rollbot 2 preserves the single-actuator design while its redesigned mass distribution enables near-straight rolling and more direct travel toward commanded paths and beacon targets. Onboard inertial and ultra-wideband sensing support learned steering, beacon seeking, and waypoint following through one motor-velocity command. Rollbot 2 autonomously approaches UWB beacons without motion capture and completes a 468~m manually commanded indoor route at 0.87~m/s. On the shared 4.8~m ``N'' waypoint benchmark, all 16 trials complete the path with a mean traversal time of 10.8~s and mean time-averaged deviation of 7~cm. Its fastest traversal takes 8.0~s, over 70 times faster than the original Rollbot's 564~s run despite minor protocol differences. These results show how mechanical redesign, onboard sensing, and learning-based control improve the agility and autonomy of a single-actuator spherical robot.
\end{abstract}


\section{Introduction}

Underactuated robots have fewer independent control inputs than mechanical degrees of freedom, offering potential savings in mass, volume, power, material, and hardware complexity \cite{urreview}. These advantages have motivated extensive work on inverted pendulums \cite{pendulumreview,quadpendulum}, robotic hands \cite{urhand1,urhand2}, and underactuated fingers and wrists \cite{urwrist1,urwrist2}. Extreme underactuation is challenging in mobile robots because planar maneuvering normally requires independent control of propulsion and steering. Single-actuator locomotion has nevertheless been demonstrated using one-way bearings \cite{zhao2013msu}, passive wheels \cite{ito2019autonomous}, and carefully timed dynamic motion in jumping \cite{zhao2013msu}, wheeled \cite{ito2019autonomous}, walking \cite{kyle2023simplest,zarrouk2012compliance}, and flying robots \cite{zhang2016controllable,curtis2023autonomous,wang2024single}. These examples show that reducing actuator count need not eliminate mobility, but shifts the burden of generating useful behavior to the mechanism, passive dynamics, and controller.

\begin{figure}[!t]
\centering
\includegraphics[width=\linewidth]{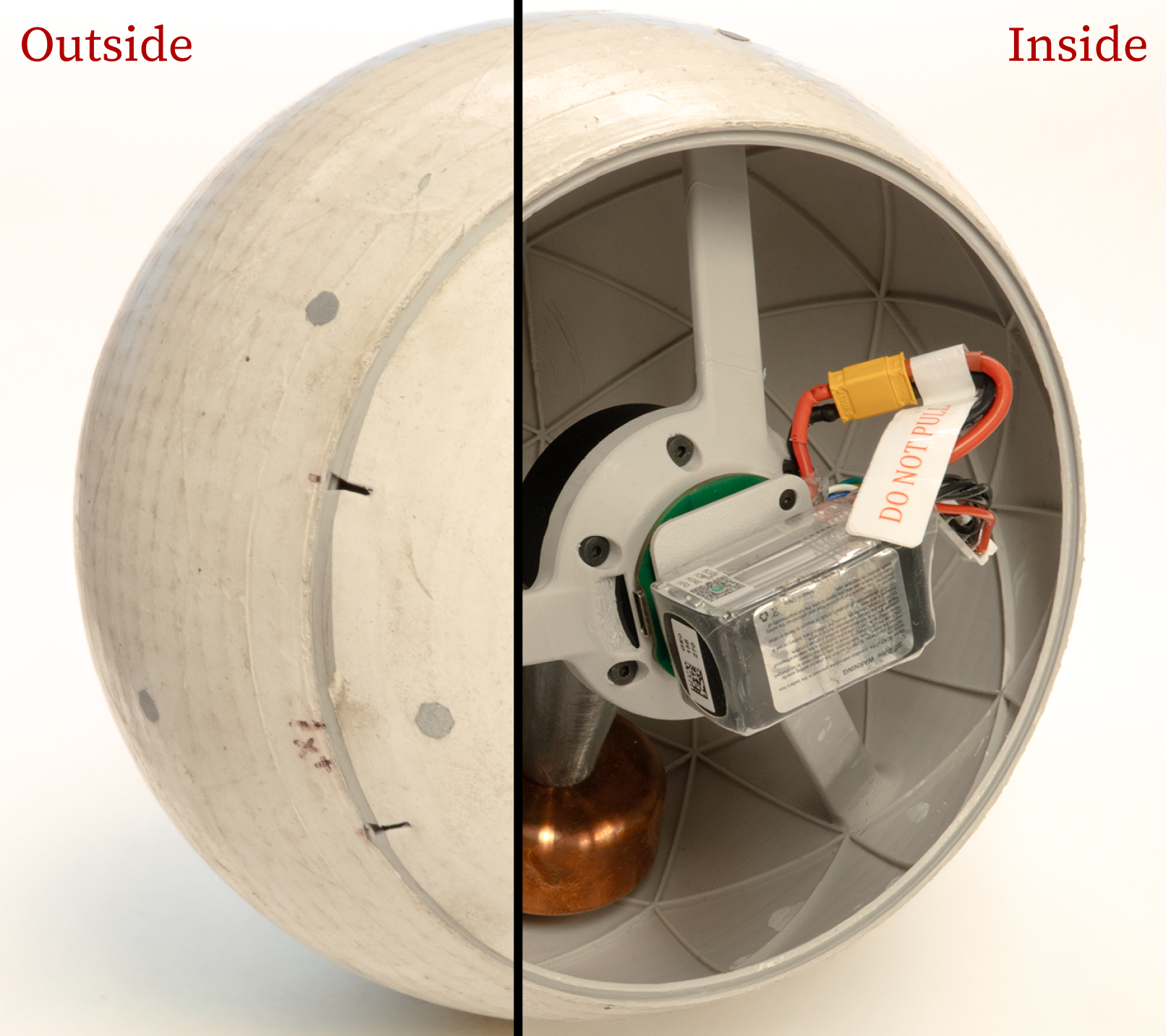}
\caption{Interior and exterior of Rollbot 2. The robot has an outer diameter of 20~cm and weighs 0.90~kg.}
\label{fig:rollbot2-hardware}
\end{figure}

Spherical robots are a compelling platform for studying this tradeoff. Their shells protect the internal mechanisms and electronics, allow rolling without exposed wheels or legs, and are attractive for operation in environments where robustness is important. Conventional spherical robots use barycentric mechanisms, control-moment gyroscopes, or deformable shells to control their motion \cite{diouf2024spherical,halme1996motion,bhattacharya2000spherical,mojabi2002introducing,wait2010self,chen2016design}. These mechanisms generally use multiple actuators or multiple independently controlled internal degrees of freedom to generate propulsion and steering. Although models have considered rolling balls driven by internal point masses \cite{putkaradze2018dynamics}, obtaining fast, controlled two-dimensional motion from one physical actuator remains difficult because the same input must propel the robot, regulate its rolling state, and redirect its trajectory.

The original Rollbot demonstrated controlled planar locomotion with a single internal motor \cite{wang2026rollbot}. It used a rotating pendulum and operated near a quasi-stable circular trajectory. Slowly accelerating and decelerating the pendulum varied trajectory curvature and guided the robot between waypoints. This proof-of-concept design established controllability but had limited speed and relied on OptiTrack for global position and orientation feedback. In work developed at about the same time, ROCK demonstrated reinforcement-learning-based steering of a one-degree-of-freedom internal-pendulum robot with an uneven shell \cite{xu2025steerable}. Its nonspherical contact surface provides steering authority, and the robot operates in an OptiTrack motion-capture environment. These studies motivate fast locomotion with a spherical shell and onboard sensing.

This paper presents Rollbot 2 as a performance-oriented redesign of the original Rollbot. The goal is not only to demonstrate controllability with one actuator, but to bring the speed and maneuverability of a mechanically minimal robot closer to those of conventional spherical robots. First, we redesign the mass distribution so that approximately constant pendulum rotation produces continuous straight-line rolling instead of a quasi-stable helical trajectory. Transient modulation of the same motor command then generates responsive steering. Second, we integrate an inertial measurement unit (IMU) and ultra-wideband (UWB) ranging module for onboard motion estimation and target-relative navigation. Third, we use reinforcement learning to exploit dynamic motions that were deliberately suppressed by the original quasi-stable controller.

We study three complementary tasks: commanded rolling and steering, onboard beacon seeking, and waypoint following. Together, they examine how mechanical redesign, onboard sensing, and learning-based control extend the capabilities of a single-actuator spherical robot.

\section{System Overview}
\label{sec:Rollbot2-overview}

Rollbot 2 preserves the core single-actuator architecture \cite{wang2026rollbot}: an internal motor drives a pendulum-like mass to produce shell motion. The redesigned mass distribution favors continuous straight rolling, while transient motor-speed changes generate steering (Fig.~\ref{fig:rollbot2-mass-distribution}).

The onboard system is also redesigned. Rollbot 2 integrates an IMU for local motion estimation and a UWB ranging module for distance measurements to external beacons. Commanded steering and beacon seeking use these sensors without OptiTrack feedback, while the waypoint benchmark retains external localization. The IMU and UWB range history allow the robot to estimate relative beacon position and approach the beacon using onboard computation and sensing.

The control architecture combines a learned policy with the motor's native velocity controller. At 50~Hz, the policy maps a short observation history and a task specification to one motor target velocity. The firmware clips this output to $\pm20\,$rad/s at the geared output and sends velocity commands to the motor controller at 200~Hz. Separate policies share this interface for commanded steering, beacon seeking, and waypoint following. The policies exploit transient and oscillatory motions that were intentionally excluded from the original Rollbot's quasi-stable controller.

\begin{figure}[htbp]
\centering
\includegraphics[width=\linewidth]{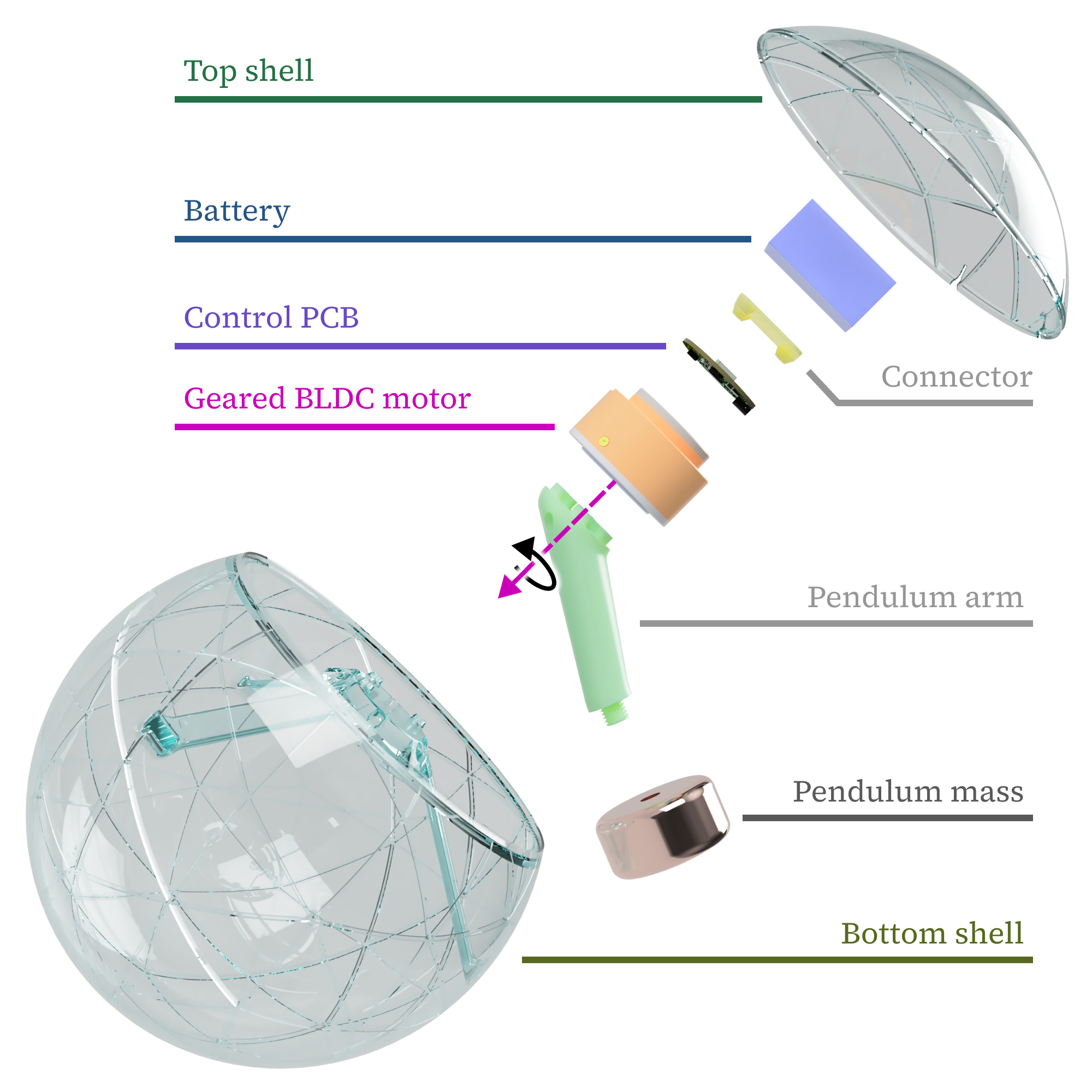}
\caption{Exploded view of Rollbot 2. As in the original Rollbot, the sole actuator is a central geared BLDC motor that drives the pendulum around the purple axis.}
\label{fig:rollbot2-exploded}
\end{figure}

\section{Hardware Redesign}
\label{sec:Rollbot2-hardware}

\begin{figure}[htbp]
\centering
\begingroup
\definecolor{illustrationgray}{RGB}{217,217,217}
\setlength{\unitlength}{\linewidth}
\begin{picture}(1,1)
\put(0,0){\includegraphics[width=\linewidth]{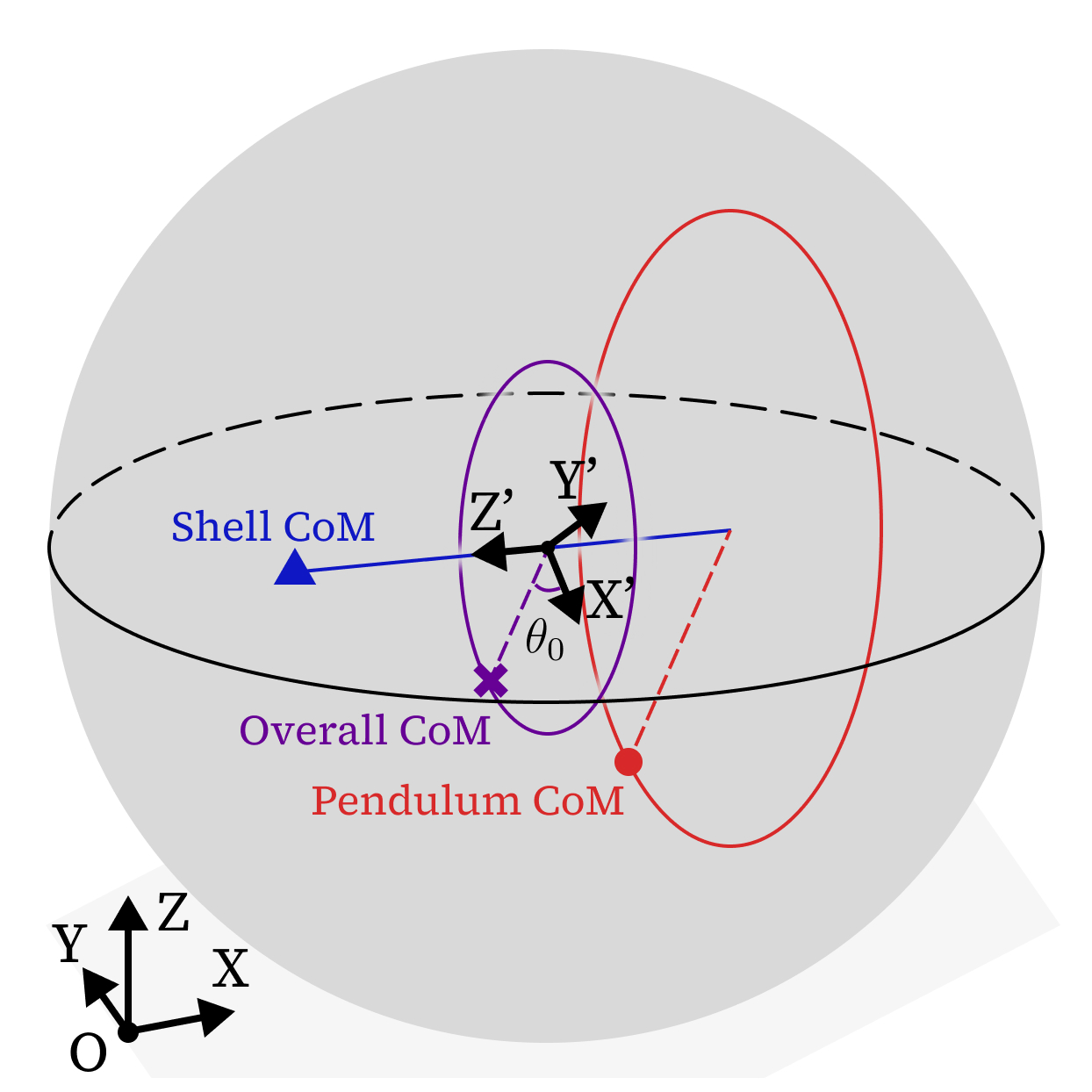}}
\put(0.474,0.391){\color{illustrationgray}\rule{0.052\linewidth}{0.050\linewidth}}
\put(0.483,0.400){\color{black}$\phi$}
\end{picture}
\endgroup
\caption{Illustration of the Rollbot 2 mass distribution. The shell CoM (blue triangle) and pendulum CoM (red circle) are both offset from the shell center, while the shell center lies approximately in the overall CoM trajectory plane. Not drawn to scale. The ground frame is denoted by $S$, and the local body frame denoted by $S'$ has its origin at the sphere center with $Z'$ axis aligned with the motor rotation axis.}
\label{fig:rollbot2-mass-distribution}
\end{figure}

The main mechanical change is the weight distribution. The original Rollbot had its shell center of mass at the sphere center and an offset internal pendulum. In Rollbot 2, the shell and pendulum centers of mass have opposing axial offsets of magnitudes $d_s$ and $d_{pz}$. With shell mass $M$ and pendulum mass $m$, their axial moments approximately balance:
\begin{equation}
    M d_s \approx m d_{pz}.
    \label{eq:rollbot2-axial-balance}
\end{equation}
Relative to the shell, the overall center of mass therefore traces a circle in a plane approximately through the sphere center, perpendicular to the motor axis (Fig.~\ref{fig:rollbot2-mass-distribution}). This removes the persistent axial mass bias associated with the original circular or helical trajectory, enabling periodic pendulum rotation to sustain approximately straight rolling.

Changing motor speed perturbs the coupled shell--pendulum motion and generates steering. The controller uses these transients to redirect the robot while maintaining rolling.

The control PCB integrates computation, motor control, inertial sensing, and UWB ranging. The IMU supplies orientation, angular velocity, and acceleration; the motor driver supplies motor angle and velocity; and the UWB module measures range to external beacons. These measurements support control outside a motion-capture arena. Figures~\ref{fig:rollbot2-hardware} and~\ref{fig:rollbot2-exploded} show the prototype and mechanical and electronics layout.

\begin{table*}[htbp]
    \centering
    \caption{Physical and actuator parameters of Rollbot 2.}
    \label{tab:rollbot2-params}
    \begin{tabularx}{\textwidth}{|c|l|X|}
        \hline
        \textbf{Quantity} & \textbf{Value} & \textbf{Explanation} \\
        \hline
        $R$ & $0.10\,$m & Outer radius of shell\\
        $M$ & $0.445\,$kg & Mass of shell\\
        $m$ & $0.455\,$kg & Pendulum mass\\
        $d_s$ & $0.0273\,$m & Shell CoM offset from shell center\\
        $d_{pz}$ & $0.0280\,$m & Pendulum CoM offset from shell center in the $Z'$ direction\\
        $r_p$ & $0.0702\,$m & Pendulum CoM offset from motor rotation axis\\
        $I_{z'z'}$ & $1.38\times10^{-3}\,\text{kg}\cdot\text{m}^2$ & Shell moment of inertia about $Z'$ through shell center\\
        $I_{x'x'}=I_{y'y'}$ & $1.93\times10^{-3}\,\text{kg}\cdot\text{m}^2$ & Shell moment of inertia about $X'$ and $Y'$ through shell center\\
        $\tau_{\max}$ & $1\,\text{N}\cdot\text{m}$ & Maximum motor torque\\
        $\omega_{\max}$ & $20\,\text{rad/s}$ & Maximum motor speed\\
        \hline
    \end{tabularx}
\end{table*}

\section{Reinforcement-Learning-Based Control}
\label{sec:Rollbot2-control}

We train separate policies for commanded steering, beacon seeking, and waypoint following using proximal policy optimization (PPO) in MuJoCo/MJX \cite{schulman2017ppo}. Each is a feedforward network with two 128-unit ReLU hidden layers. Its input stacks three observations $\{o_{t-2},o_{t-1},o_t\}$ containing proprioception, previous action, and task variables. This short history provides temporal context under partial observability \cite{hausknecht2015deep}. The output is a scalar motor target velocity clipped to $\pm20\,$rad/s, with control period $\Delta t_{\mathrm{ctrl}}=0.02\,$s and simulation step $\Delta t_{\mathrm{sim}}=0.002\,$s.

In simulation, a proportional--integral velocity loop converts tracking error into torque, with $K_p=2.0\,\text{N}\cdot\text{m}\cdot\text{s/rad}$, $K_i=10\,\text{N}\cdot\text{m/rad}$, and motor-velocity damping $0.005\,\text{N}\cdot\text{m}\cdot\text{s/rad}$. The torque limit decreases linearly from $1\,\text{N}\cdot\text{m}$ at rest to zero at $20\,$rad/s. Hardware uses the motor's native velocity controller.

For each task, the reward accumulated over one control step is

\begin{equation}
    r_t = \Delta t_{\mathrm{ctrl}}\,\rho_t,\qquad
    \rho_t = \sum_i w_i r_{i,t},
    \label{eqn:rollbot2-rl-reward}
\end{equation}

where $\rho_t$ combines task progress and regularization terms evaluated numerically in SI units. We denote motor-velocity output by $a_t$ and planar translational velocity by $\vec{v}_t$. Rewards use true simulator state. The steering actor and critic share 36 inputs; the beacon critic adds two true target-frame velocities (35 inputs), and the waypoint critic adds four true path coordinates and velocities (55 inputs). These additional critic inputs are used only during training.

The observations remove absolute configuration variables that do not change the required behavior. For steering, rotating both the robot and its commanded direction by the same angle should leave the motor command unchanged: moving north or east is the same control problem when the robot's orientation relative to the command is identical. We therefore express orientation and angular velocity in a command-aligned frame. Similarly, consider two configurations that differ only by rotating the shell about the motor axis while keeping the pendulum fixed. Both the body-frame orientation and motor encoder angle change, but the physical mass configuration remains identical. Instead of using a quaternion and raw motor angle, we therefore describe orientation by the motor-axis direction and motor phase relative to gravity. Both remain unchanged under this shell rotation, so equivalent configurations produce the same observation.

The shared inputs are motor-axis direction $\hat{z}'_t$, shell angular velocity $\vec{\omega}_t$, gravity-relative motor phase $m_t$, motor velocity $\dot{\phi}_t$, and previous clipped command $a_{t-1}$. We denote motor angle by $\phi_t$ and motor angular velocity by $\dot{\phi}_t$. Empirical input normalization is folded into the first layer's weights and bias for deployment.

We use the left-superscript convention ${}^{F}a$ to denote that quantity $a$ is expressed in frame $F$. Translational velocities are planar, while orientation axes and angular velocities are three-dimensional. In cross products, planar velocity is embedded in three dimensions with a zero vertical component.

\subsection{Commanded Rolling and Steering Policy}
\label{sec:Rollbot2-rl-steering}

The commanded-motion policy tracks a desired planar travel direction $\hat{x}^{\mathrm{cmd}}_t=[\cos\psi_t^{\mathrm{cmd}},\sin\psi_t^{\mathrm{cmd}}]^T$ and speed $v^{\mathrm{cmd}}_t$, with direction-change rate $\omega^{\mathrm{cmd}}_t=\dot{\psi}_t^{\mathrm{cmd}}$. Training varies speed between 0.1 and 1.6~m/s and turning rate within $\pm0.7\,$rad/s, exposing the policy to straight motion, speed changes, and turns. On the robot, joystick magnitude sets speed over an active range of 0.2--1.2~m/s, while joystick direction specifies heading. A first-order heading filter with a 0.8~s time constant smooths direction changes, and the smoothed angular rate supplies $\omega^{\mathrm{cmd}}_t$.

The command-aligned frame $S''$ has $\hat{x}''$ along $\hat{x}^{\mathrm{cmd}}_t$, $\hat{z}''$ vertical, and $\hat{y}''=-\hat{x}''\times\hat{z}''$, removing arbitrary global heading from the observation.

For one time step, the policy observation is
\begin{equation}
    o_t = [\,{}^{S''}\hat{z}'_t,\ {}^{S''}\vec{\omega}_t,\ m_t,\ \dot{\phi}_t,\ a_{t-1},\ v^{\mathrm{cmd}}_t,\ \omega^{\mathrm{cmd}}_t\,] \in \mathbb{R}^{12},
\end{equation}
Here $m_t\in\mathbb{R}^2$, and the two command inputs specify desired speed and frame turning rate. Translational velocity is used in the reward but omitted from the actor input. The three-frame policy input is $36$-dimensional.

We compute $m_t$ by expressing the motor-frame unit vector $[\cos \phi_t,\sin \phi_t]^T$ relative to the projection of gravity in the body $X'Y'$ plane. The relative phase $\zeta_t$ is zero at the gravity-aligned bottom position, and gravity-induced torque varies with this phase:

\begin{equation}
\begin{aligned}
    m_t &= \begin{bmatrix}\cos\zeta_t \\ \sin\zeta_t\end{bmatrix}
    = \operatorname{Rot}(g_{x'y'})
    \begin{bmatrix}\cos \phi_t \\ \sin \phi_t\end{bmatrix}, \\
    \operatorname{Rot}(g_{x'y'}) &=
    \frac{\begin{bmatrix}
        g_{x'} & g_{y'} \\
        -g_{y'} & g_{x'}
    \end{bmatrix}}{\sqrt{g_{x'}^2+g_{y'}^2}}.
\end{aligned}
\end{equation}

A small denominator regularizer handles nearly vertical motor axes. This encoding retains pendulum phase relative to gravity without requiring absolute rotation about the motor axis.

The reward rate tracks the commanded speed along the desired direction while suppressing sideways drift, slip, unfavorable body pose, and abrupt action changes:
\begin{equation}
\begin{split}
    \rho_t ={}& r^{x}_t - w_y\alpha_t r^{y}_t
    - w_{\mathrm{slip}}\alpha_t r^{\mathrm{slip}}_t \\
    &- w_a\alpha_t r^{\Delta a}_t - w_{\mathrm{pose}} r^{\mathrm{pose}}_t,
\end{split}
\end{equation}
with
\begin{align}
    r^{x}_t &= b_x - \lvert \vec{v}_t\cdot\hat{x}^{\mathrm{cmd}}_t - v^{\mathrm{cmd}}_t \rvert, \\
    r^{y}_t &= \left\|\vec{v}_t - (\vec{v}_t\cdot\hat{x}^{\mathrm{cmd}}_t)\hat{x}^{\mathrm{cmd}}_t\right\|, \\
    r^{\mathrm{slip}}_t &= \left\|\vec{\omega}_t\times[0,0,R]^T - \vec{v}_t\right\|, \\
    r^{\Delta a}_t &= \left\|a_t-a_{t-1}\right\|^2, \\
    r^{\mathrm{pose}}_t &= \left\|\hat{z}'_t - (\hat{z}'_t\cdot\hat{y}^{\mathrm{cmd}}_t)\hat{y}^{\mathrm{cmd}}_t\right\|, \\
    \hat{y}^{\mathrm{cmd}}_t &= [-\sin\psi_t^{\mathrm{cmd}},\ \cos\psi_t^{\mathrm{cmd}},\ 0]^T.
\end{align}
Here, $\lvert\cdot\rvert$ denotes scalar absolute value and $\|\cdot\|$ the Euclidean norm. The longitudinal term $r^x_t$ rewards speed tracking; the lateral term $r^y_t$ discourages sideways drift. The slip term $r^{\mathrm{slip}}_t$ measures disagreement between shell rotation and translation under rolling without slip. The action-change term $r^{\Delta a}_t$ discourages abrupt commands, while the pose term $r^{\mathrm{pose}}_t$ favors a horizontal motor axis perpendicular to travel.

The speed-tracking offset is $b_x=2$. The lateral, slip, action-change, and pose weights are $w_y=1.5$, $w_{\mathrm{slip}}=0.3$, $w_a=0.05$, and $w_{\mathrm{pose}}=0.5$, respectively; the latter three are reused below.

The curriculum uses $\alpha_t=\alpha_0+(1-\alpha_0)p_t$, with initial penalty fraction $\alpha_0=0.1$ and $p_t\in[0,1]$ measuring training progress over the first 3000 iterations at the default batch size. Gradually increasing the lateral, slip, and action-change penalties lets the policy discover locomotion before refining gait quality.

To reduce sensitivity to simulation error, steering-policy training randomizes translational and rolling friction, joint friction, shell and pendulum mass properties, initial state, external force and torque disturbances, and observation noise.

\subsection{Beacon-Seeking Policy}
\label{sec:Rollbot2-rl-beacon}

The beacon-seeking task couples onboard estimation with dynamic locomotion. UWB provides range but not bearing, so an extended Kalman filter (EKF) combines range history with robot motion to estimate relative beacon position. The filter operates in polar coordinates $(r,\xi)$, separating range and bearing uncertainty. The policy uses an estimated-target-aligned frame $\hat{B}$ whose $x$ axis points towards the estimated target position and whose $z$ axis is vertical.

Assuming rolling without slip and a stationary beacon, the estimator predicts relative beacon velocity as $\vec{v}_{b}=-\vec{\omega}\times R\hat{z}$. At approximately 100~Hz, it integrates IMU-derived motion and corrects the estimate with timestamped UWB ranges.

During training, perturbed target position and proprioceptive observations expose the policy to estimation error; the firmware EKF is used at deployment.

For one time step, the policy observation is
\begin{equation}
    o_t = [\,\tilde{d}_{b,t},\ {}^{\hat{B}}\hat{z}'_t,\ {}^{\hat{B}}\vec{\omega}_t,\ m_t,\ \dot{\phi}_t,\ a_{t-1}\,] \in \mathbb{R}^{11},
    \label{obs:target}
\end{equation}

where $\tilde{d}_{b,t}=A_b\tanh(\hat{d}_{b,t}/\ell_b)$ encodes estimated target distance $\hat{d}_{b,t}$ as a dimensionless, bounded input, with saturation amplitude $A_b=2$ and distance scale $\ell_b=2~\mathrm{m}$. The target-aligned frame removes absolute target position and global heading: rotating or translating the robot and target together leaves the relative control problem unchanged. Bounding distance adds an approximate far-field invariance, since targets along the same bearing should call for similar approach behavior once they are sufficiently distant. Close to the target, the encoding retains distance information needed for slowing and arrival. The three-frame policy input is $33$-dimensional.

For rewards, $d_{b,t}$ denotes true target distance and $B$ is aligned with the true target direction. The reward rate encourages approach, arrival, and reduced speed near the goal:
\begin{equation}
\begin{split}
    \rho_t ={}& r^{d}_t + w_g r^{g}_t + r^{v}_t \\
    &+ w_{\mathrm{slow}} r^{\mathrm{slow}}_t - w_a r^{\Delta a}_t,
\end{split}
\end{equation}
with
\begin{align}
    r^{d}_t &= b_d - \frac{d_{b,t}}{\ell_d}, \\
    r^{g}_t &= \exp\!\left(-\frac{d_{b,t}^2}{\ell_g^2}\right), \\
    r^{v}_t &= \vec{v}_t\cdot\hat{x}^{B}_t - k_v\max\left(\|\vec{v}_t\|-v_{\mathrm{lim}}^b,\ 0\right), \\
    r^{\mathrm{slow}}_t &= -\|\vec{v}_t\|\exp\!\left(-\frac{d_{b,t}^2}{\ell_{\mathrm{slow}}^2}\right).
\end{align}
The reward uses true target geometry, while the actor receives the perturbed estimate. The distance term $r^d_t$ has offset $b_d=1$ and decay scale $\ell_d=5~\mathrm{m}$. The projected-velocity term $r^v_t$ rewards approach and penalizes speeds above $v_{\mathrm{lim}}^b=1~\mathrm{m/s}$ with gain $k_v=2$, reused for waypoint following. The arrival term $r^g_t$ has weight $w_g=2$ and scale $\ell_g=0.2~\mathrm{m}$, while $r^{\mathrm{slow}}_t$ has weight $w_{\mathrm{slow}}=0.3$ and a narrower scale $\ell_{\mathrm{slow}}=0.1~\mathrm{m}$, encouraging fast approach followed by deceleration near the goal. Omitting separate lateral, slip, and pose penalties leaves freedom to redirect motion toward the beacon. The action-change penalty $r^{\Delta a}_t$ follows Section~\ref{sec:Rollbot2-rl-steering}; beacon and waypoint training use the raw current output and previous clipped command, while steering clips both.

\subsection{Waypoint-Following Policy}
\label{sec:Rollbot2-rl-waypoint}

The waypoint controller learns local transitions between two segments separated by turn angle $\psi^{\mathrm{turn}}$, with randomized geometry and initial state. At deployment, expressing the robot state relative to each successive turn composes these local behaviors into longer paths, including the ``N'' trajectory.

\begin{figure}[htbp]
\centering
\includegraphics[width=\linewidth]{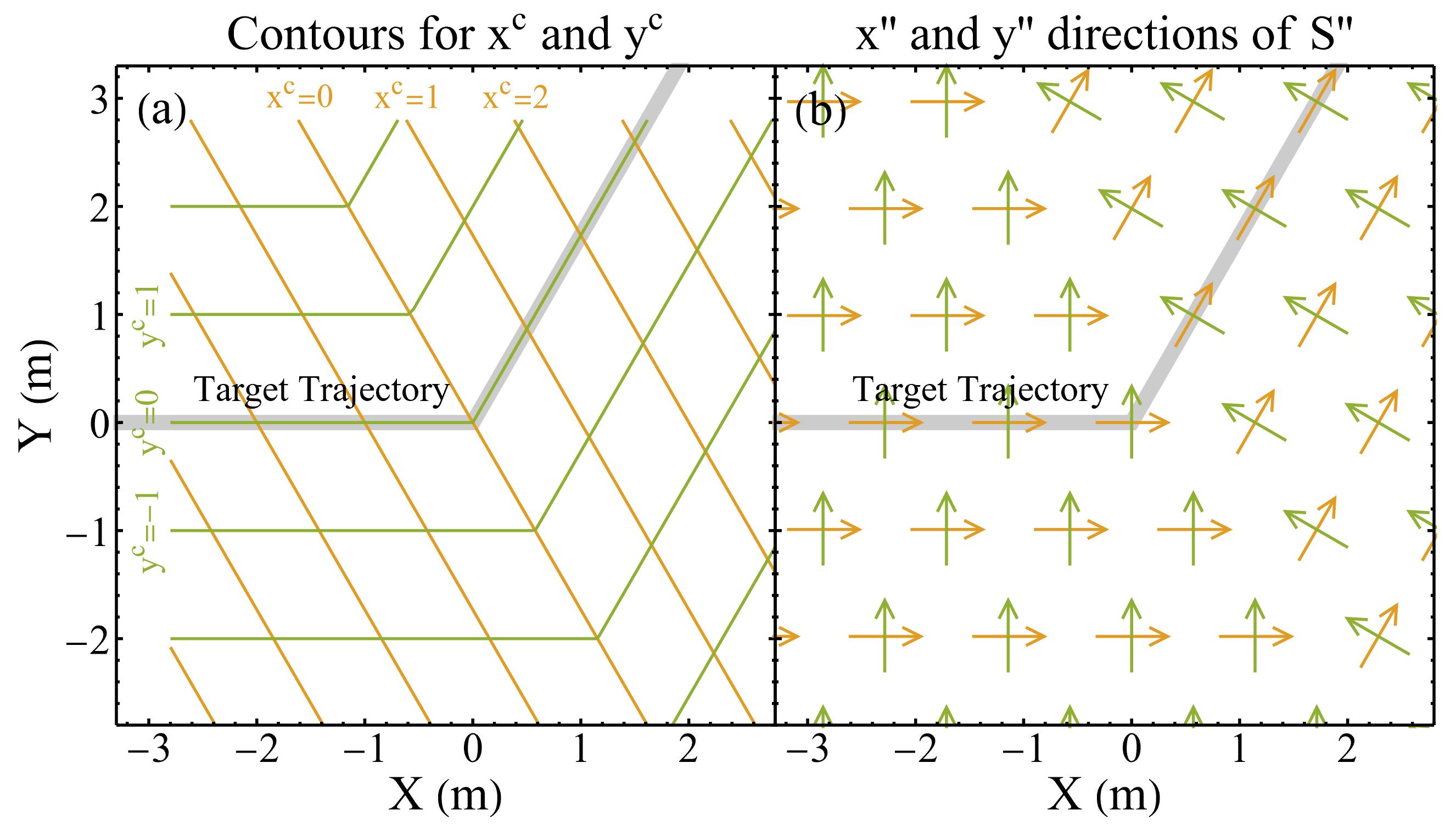}
\caption{Waypoint coordinate systems for a $60^\circ$ turn. (a) Contours of progress $x^c$ (orange) and cross-track error $y^c$ (green). (b) Active path tangent $x''$ (orange) and left-normal direction $y''$ (green), with $\hat{y}''=-\hat{x}''\times\hat{z}''$.}
\label{fig:rollbot2-path-coords}
\end{figure}

Generalized coordinates $(x^c,y^c)$ describe progress through a turn and signed cross-track deviation (Fig.~\ref{fig:rollbot2-path-coords}(a)). Intuitively, the green contours form a family of paths parallel to the desired trajectory: $y^c=0$ is the desired path, and nonzero $y^c$ shifts it to the left or right. Along each such path, $x^c$ measures signed travel distance from its own corner, with negative values before the turn, zero at the corner, and positive values afterward. This gives the robot a continuous description of how far it has progressed and how far it has drifted, even as the path changes direction.

To define these coordinates precisely, place the desired corner at the origin of an incoming-aligned plane, with planar position $\vec{p}=[X,Y]^T$ and signed turn angle $\psi^{\mathrm{turn}}$. The incoming and outgoing tangents and left normals are
\begin{align}
    \hat{e}_0&=[1,0]^T, & \hat{n}_0&=[0,1]^T, \\
    \hat{e}_1&=[\cos\psi^{\mathrm{turn}},\sin\psi^{\mathrm{turn}}]^T, &
    \hat{n}_1&=[-\sin\psi^{\mathrm{turn}},\cos\psi^{\mathrm{turn}}]^T.
\end{align}
Writing $s_i=\vec{p}\cdot\hat{e}_i$, $d_i=\vec{p}\cdot\hat{n}_i$, and $\kappa=\tan(\psi^{\mathrm{turn}}/2)$, the continuous turn coordinates are
\begin{equation}
    (x^c,y^c)=
    \begin{cases}
        (s_0+\kappa d_0,\ d_0), & s_0+s_1\leq0, \\
        (s_1-\kappa d_1,\ d_1), & s_0+s_1>0.
    \end{cases}
    \label{eq:rollbot2-turn-coordinates}
\end{equation}
The bisector $s_0+s_1=0$ selects the active segment $i$. Its frame $S''$ has $\hat{x}''$ along $\hat{e}_i$, vertical $\hat{z}''$, and $\hat{y}''=-\hat{x}''\times\hat{z}''$. Projected velocities are $v^c_x=\vec{v}\cdot\hat{e}_i$ and $v^c_y=\vec{v}\cdot\hat{n}_i$. Within each region, $\dot{x}^c=v^c_x+\kappa v^c_y$ on entry and $\dot{x}^c=v^c_x-\kappa v^c_y$ on exit, explaining why progress rate differs from tangential velocity.

For one time step, the policy observation is
\begin{equation}
\begin{split}
    o_t = [\,&\cos\tilde{x}^c_t,\ \sin\tilde{x}^c_t,\ y^c_t,
    \ v^c_{x,t},\ v^c_{y,t}, \\
    &{}^{S''}\vec{\omega}_t,\ {}^{S''}\hat{z}'_t,\ m_t,
    \ \dot{\phi}_t,\ a_{t-1}, \\
    &\cos\psi^{\mathrm{turn}},\ \sin\psi^{\mathrm{turn}}\,]
    \in \mathbb{R}^{17},
\end{split}
\end{equation}
The bounded encoding $\tilde{x}^c_t=\pi\tanh(x^c_t/\ell_x)$, with progress scale $\ell_x=1~\mathrm{m}$, distinguishes approach and departure near the corner but converges to the same straight-tracking representation far on either side. Cross-track error and velocity describe displacement and drift; the sine and cosine of $\psi^{\mathrm{turn}}$ specify turn geometry continuously. The three-frame policy input is $51$-dimensional.

The reward rate encourages forward motion while penalizing cross-track error, excessive speed, and irregular actuation:
\begin{equation}
\begin{split}
    \rho_t ={}& r^{x^c}_t - \beta_t r^{y^c}_t - r^{\mathrm{vel}}_t
    - w_{\mathrm{slip}}\alpha_t r^{\mathrm{slip}}_t \\
    &- w_a\alpha_t r^{\Delta a}_t - w_{\mathrm{pose}} r^{\mathrm{pose}}_t,
\end{split}
\end{equation}
with
\begin{align}
    r^{x^c}_t &= v^c_{x,t}, \\
    r^{y^c}_t &= \left(\frac{y^c_t}{\ell_y}\right)^2, \\
    r^{\mathrm{vel}}_t &= k_v\max\!\left(\|\vec{v}_t\|-v_{\mathrm{lim}}^c,\ 0\right), \\
    r^{\mathrm{pose}}_t &= \left\|\hat{z}'_t-(\hat{z}'_t\cdot\hat{y}''_t)\hat{y}''_t\right\|.
\end{align}
The uncapped forward term $r^{x^c}_t$ promotes progress, while the quadratic path term $r^{y^c}_t$ discourages deviation with cross-track scale $\ell_y=0.20~\mathrm{m}$. The velocity term $r^{\mathrm{vel}}_t$ penalizes planar speed above $v_{\mathrm{lim}}^c=0.8~\mathrm{m/s}$ using the shared gain $k_v$. The slip term $r^{\mathrm{slip}}_t$ and action-change term $r^{\Delta a}_t$ retain their earlier roles and weights, while $r^{\mathrm{pose}}_t$ aligns the motor axis with the active segment's lateral direction. Using the same training progress $p_t$ and regularization curriculum $\alpha_t$, the path curriculum is $\beta_t=\beta_0+(1-\beta_0)p_t$, with initial fraction $\beta_0=0.25$. The stronger initial path penalty keeps early turning behaviors near the desired path before accuracy and smoothness are refined.

On the robot, OptiTrack state is transformed into coordinates relative to the active corner. Firmware handles final stopping with a 0.5~s motor-command ramp, separate from the learned local-turn behavior.

\section{Experimental Results}
\label{sec:Rollbot2-experiments}

We evaluate Rollbot 2 in three tasks: commanded steering, autonomous beacon seeking, and waypoint following on an ``N'' trajectory. The first two use onboard sensing for control; the waypoint benchmark uses OptiTrack to isolate locomotion and path tracking from localization error.

\subsection{Rolling with Steering}
\label{sec:Rollbot2-steering}

\begin{figure*}[htbp]
\centering
\begin{minipage}{0.42\textwidth}
\includegraphics[width=\linewidth]{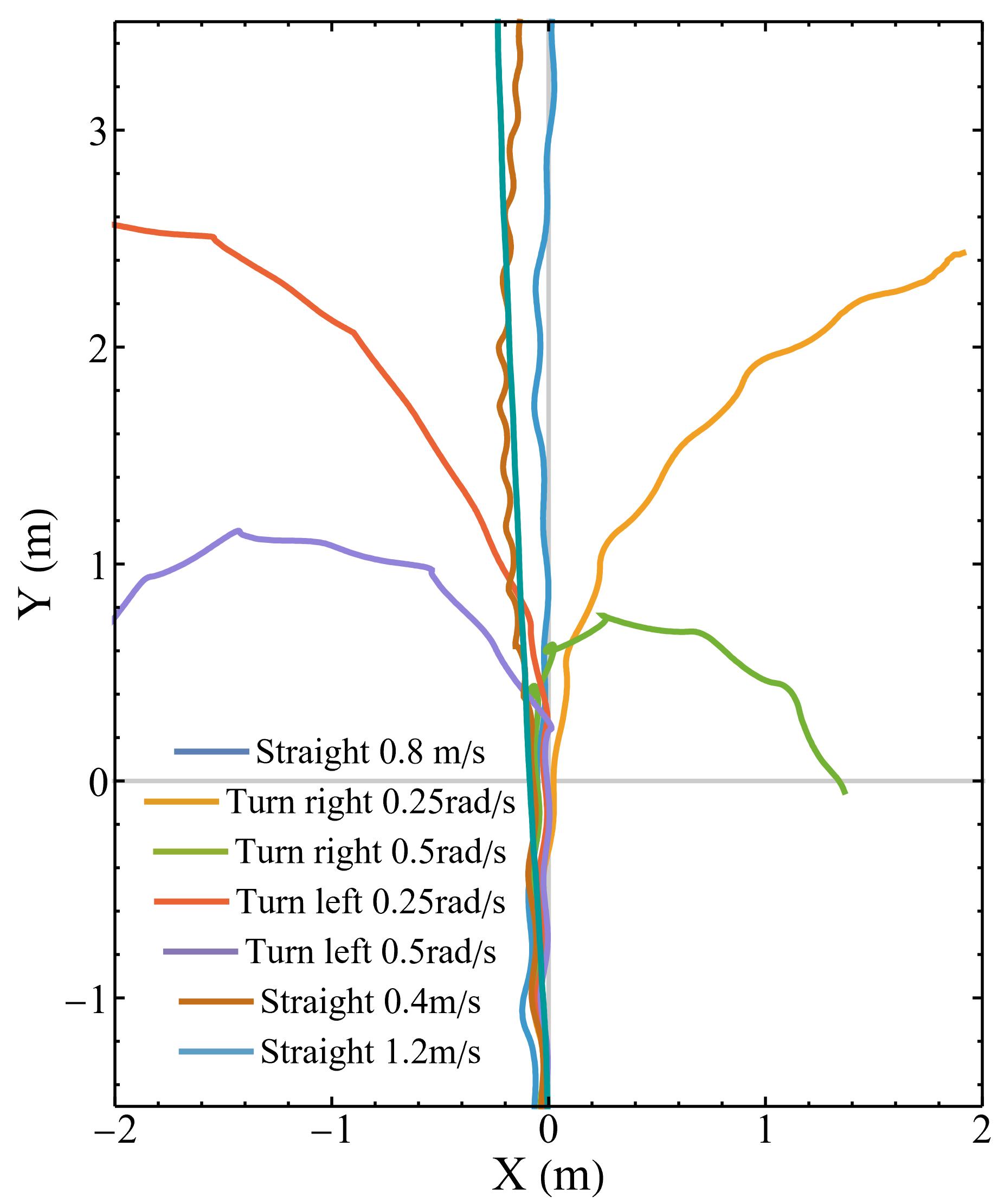}
\end{minipage}\hfill
\begin{minipage}{0.52\textwidth}
\includegraphics[width=\linewidth]{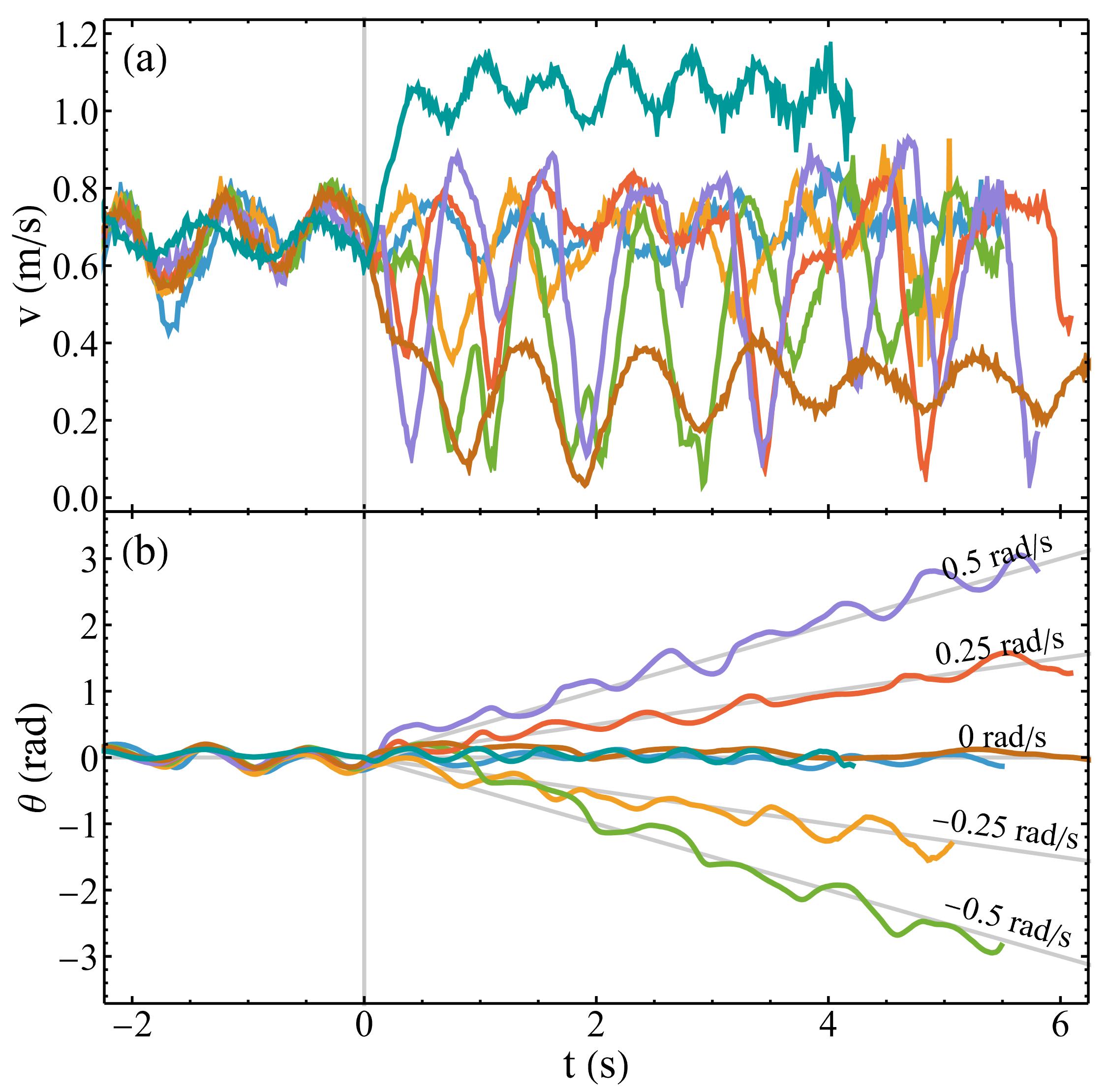}
\end{minipage}
\caption{Rolling-with-steering results. Left: planar trajectories for different steering commands. Right: measured speed and heading $\theta$ over time for the same trials.}
\label{fig:rollbot2-steering-results}
\end{figure*}

In the rolling-with-steering experiment, the policy receives desired forward speed and heading-change rate while controlling only motor target velocity. Figure~\ref{fig:rollbot2-steering-results} shows that the trajectories change direction consistently with the command sign and magnitude, while speed varies during acceleration, deceleration, and turning transients. This variation reflects the coupled nature of single-actuator control: the motor motion used to redirect the shell also affects propulsion.

In a longer manual test, we used the same controller to drive Rollbot 2 through 468~m of corridors in the Tech building, including eight right-angle turns and several smaller turns. Rollbot 2 completed the route in 9~min without hitting the walls and stayed within 30~cm of the corridor centerline for most of the trial. The corresponding average speed was 0.87~m/s and average electrical power consumption was 2.1~W.

Qualitative tests indicate that the steering policy transfers across smooth floors, slightly tilted or grainy gravel, soft mats, and relatively firm soil. Performance degrades on surfaces that are highly uneven, steep, or deformable, where the rolling-without-slip assumption and simulated contact condition become inaccurate. 

\subsection{Rolling Toward a Beacon}
\label{sec:Rollbot2-beacon}

\begin{figure*}[htbp]
\centering
\begin{minipage}{0.48\textwidth}
\includegraphics[width=\linewidth]{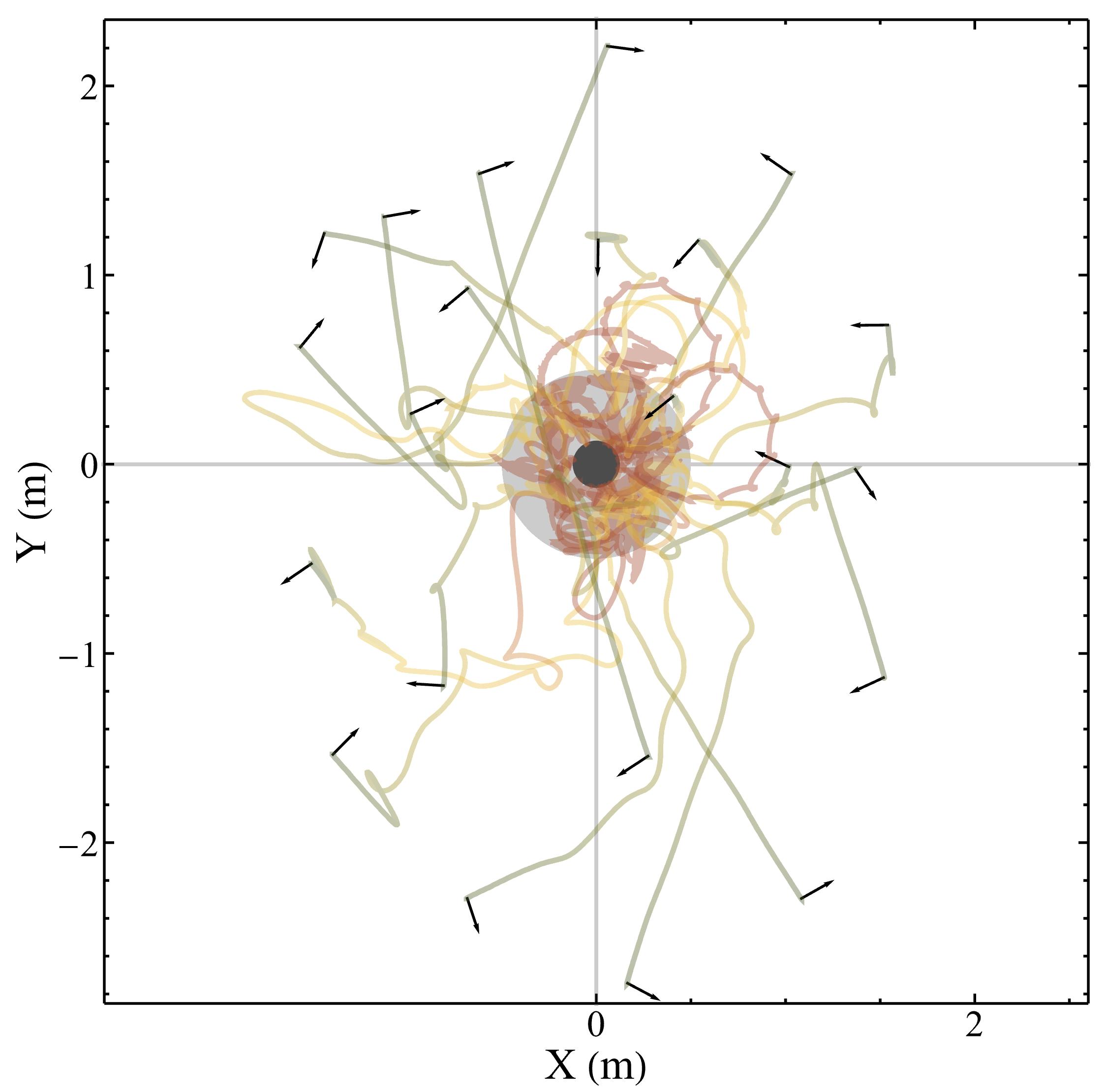}
\end{minipage}\hfill
\begin{minipage}{0.48\textwidth}
\includegraphics[width=\linewidth]{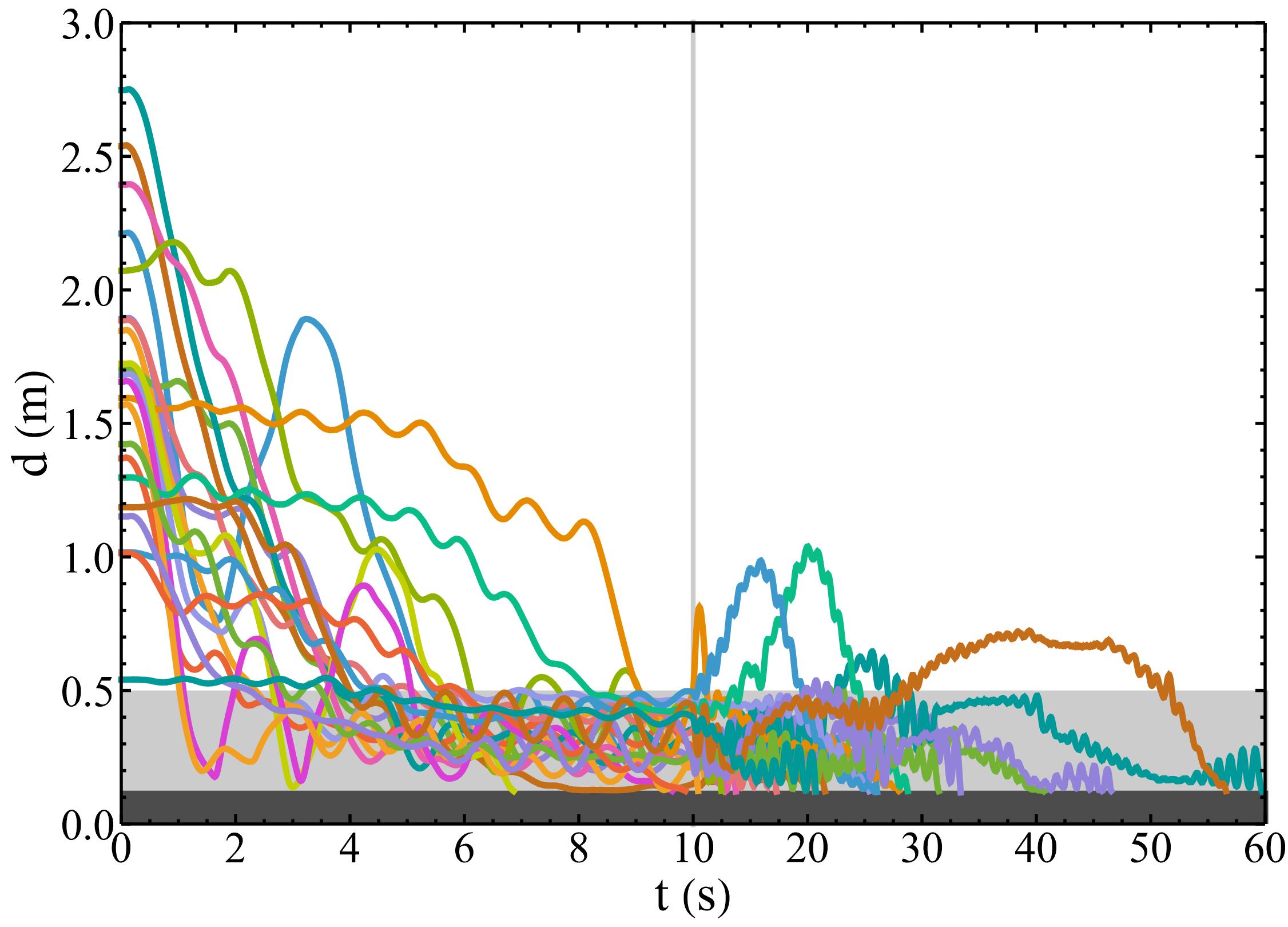}
\end{minipage}
\caption{Beacon-seeking results. Left: planar trajectories for repeated trials toward the target region. Black arrows indicate the initial $z'$ direction, and trajectory color indicates elapsed time (green: start; red: over 10~s). Right: target-distance histories for the same trials; the horizontal time axis is compressed after 10~s.}
\label{fig:rollbot2-target-results}
\end{figure*}

The beacon-seeking experiment uses no motion-capture feedback. The estimator infers target direction from IMU-derived motion and UWB range history, and the policy in Section~\ref{sec:Rollbot2-rl-beacon} converts this estimate into motor commands. Figure~\ref{fig:rollbot2-target-results} shows repeated approaches from different initial conditions. The robot generally enters the target vicinity, defined as a shell-center distance below 0.50~m, within 6~s. A target hit, defined as a distance below 0.125~m, is less consistent. UWB multipath and structural occlusion distort ranging, while residual antenna-offset and rolling-prediction errors become more consequential near the beacon. Bearing errors rotate the policy's target frame and can contribute to the observed local circling.

At larger distances, a coarse bearing estimate sustains useful progress. Near the target, the same absolute estimation error becomes a larger fraction of the remaining distance and interferes with steering and localized slowing. Rollbot 2 thus achieves fast onboard beacon approach, with precision stopping limited by near-target estimation.

Outdoor qualitative trials provided more maneuvering space and generally cleaner UWB ranging than the indoor environment. Under these conditions, Rollbot 2 typically hit targets initially 5--10~m away within 10--20~s, despite training distances being limited to 6~m. The bounded distance observation in (\ref{obs:target}) is consistent with this transfer: far-field inputs saturate, so approach behavior can depend mainly on estimated bearing. The outdoor results were not collected as a controlled statistical dataset and therefore complement the repeated indoor trajectories in Fig.~\ref{fig:rollbot2-target-results}.

\subsection{N-Shape Waypoint Following}
\label{sec:Rollbot2-n-shape}

The waypoint experiment isolates locomotion and path-following performance by using OptiTrack for accurate global state, as in the original Rollbot benchmark. Traversal time is measured from motion onset to completion of the final segment. For a trial of duration $T$, time-averaged path deviation is
\begin{equation}
    \bar{e}=\frac{1}{T}\int_0^T\min_{\vec{q}\in\mathcal{P}}
    \|\vec{p}(t)-\vec{q}\|\,dt,
    \label{eq:rollbot2-path-error}
\end{equation}
where $\vec{p}(t)$ is the robot's planar position and $\mathcal{P}$ is the entire desired polyline.

\begin{figure}[htbp]
\centering
\includegraphics[width=\linewidth]{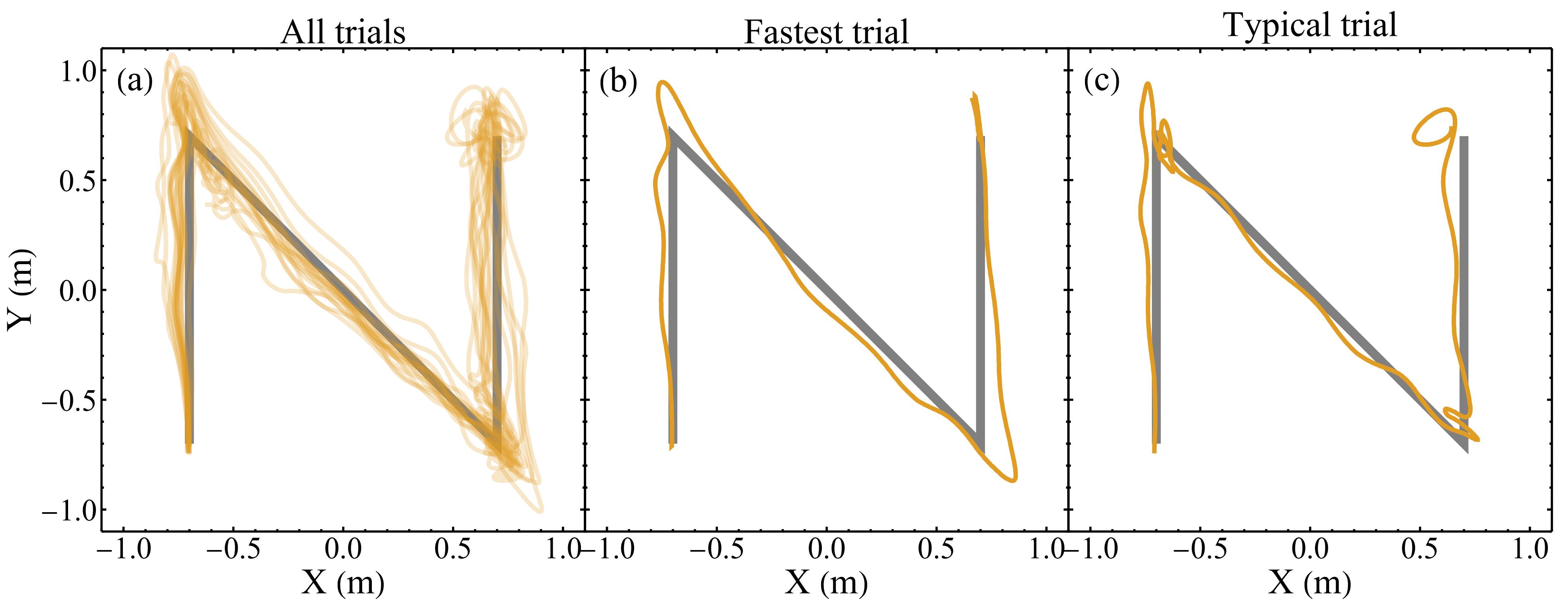}
\caption{N-shape waypoint-following results for Rollbot 2. (a) Trajectories from 16 trials following the 4.8~m ``N'' trajectory. (b) Fastest trial, which completes the path in 8.0~s. (c) A representative trial. Rollbot 2 consistently traverses the full path, although it sometimes moves back and forth near turns to adjust its heading before traversing the next line segment.}
\label{fig:rollbot2-n-shape-results}
\end{figure}

Rollbot 2 traverses the 4.8~m ``N'' path through repeated straight segments and learned turning transitions (Fig.~\ref{fig:rollbot2-n-shape-results}). All 16 trials complete the path. Mean traversal time is 10.8~s and mean time-averaged path deviation is 7~cm. The fastest trial takes 8.0~s, corresponding to desired path length divided by traversal time of 0.60~m/s. Some trials include brief reversals near a corner to adjust for heading before the robot commits to the next segment.

\begin{figure}[htbp]
\centering
\includegraphics[width=\linewidth]{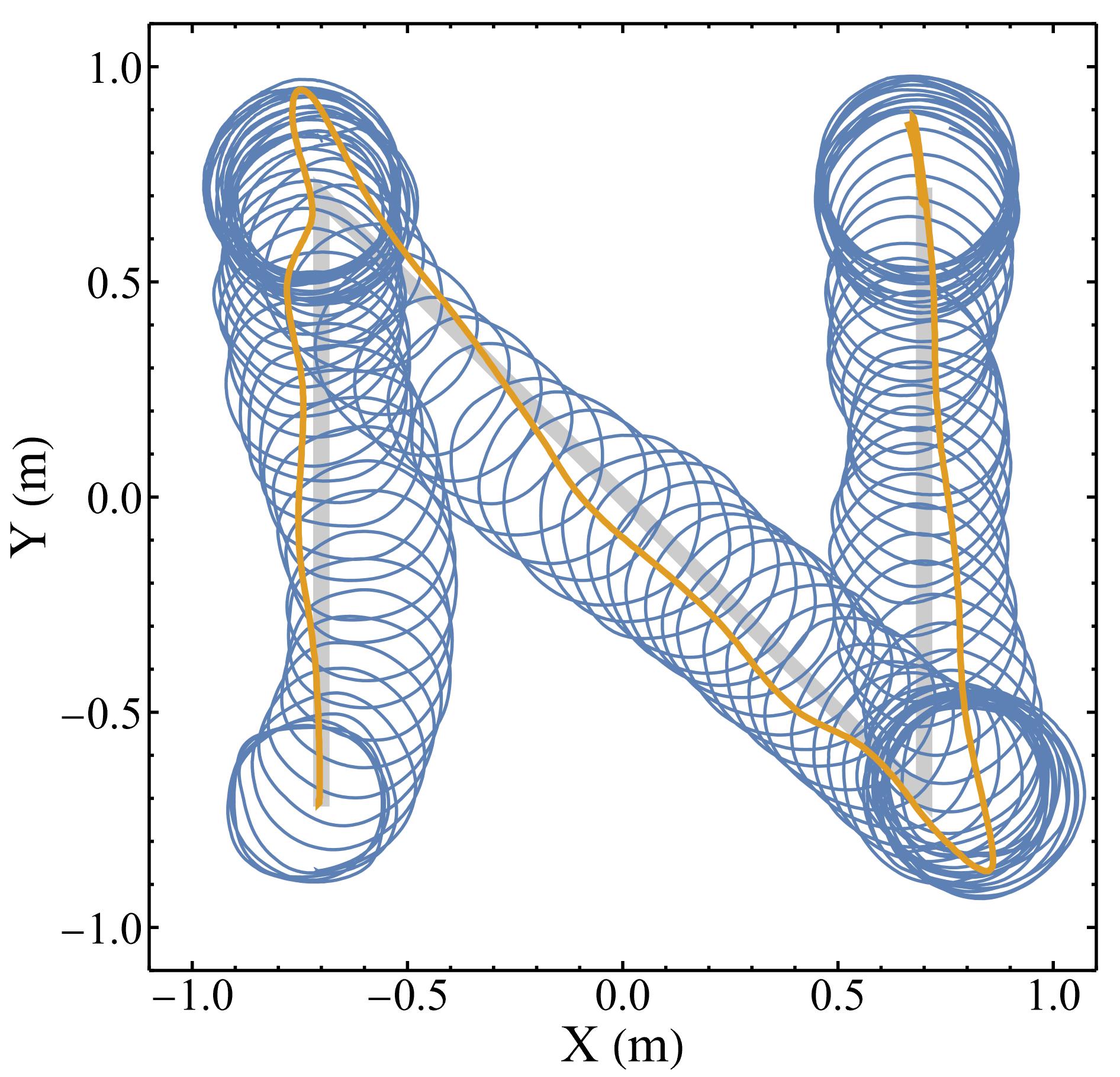}
\caption{Comparison between the original Rollbot and Rollbot 2 on the 4.8~m ``N'' trajectory. The gray line shows the desired trajectory, while the blue and orange curves show the trajectories of Rollbot 1 and Rollbot 2, respectively.}
\label{fig:rollbot2-n-shape-comparison}
\end{figure}

The original Rollbot required 564~s and had a 14~cm time-averaged deviation on the same path geometry \cite{wang2026rollbot}. Rollbot 2 reduces traversal time by approximately 52-fold on average and more than 70-fold in its fastest trial, while halving mean deviation (Fig.~\ref{fig:rollbot2-n-shape-comparison}). The original demonstration included stops at intermediate waypoints, whereas Rollbot 2 follows the path continuously. With this protocol difference in mind, the comparison shows a substantial system-level improvement from mechanical redesign and learning-based control.

\section{Discussion}
\label{sec:Rollbot2-discussion}

Rollbot 2 demonstrates that the original platform's low speed and dependence on motion capture were consequences of a particular mechanical and control design, rather than fundamental consequences of single-actuator spherical locomotion. The first Rollbot intentionally stayed near an analytically tractable quasi-stable orbit. Rollbot 2 changes the passive dynamics so that straight rolling is natural, then uses learning-based control to exploit the resulting transients. The more than 70-fold fastest-trial improvement on the common ``N'' path therefore represents a system-level gain produced by co-design.

Symmetry-aware observations are another important design choice. Expressing state relative to commands, targets, and local paths maps equivalent configurations to a shared representation, removing global pose variables that do not change the control problem. This improves data reuse across positions and headings and enables the waypoint controller to compose learned local turns into longer paths. The same principle extends to other underactuated robots whose absolute configurations contain symmetries irrelevant to task execution.

Despite these capabilities, the UWB-based beacon estimator assumes approximately rolling-without-slip motion, and performance across different surfaces has been evaluated qualitatively rather than through quantitative terrain tests. Future work will integrate additional sensors, such as cameras, to expand the robot's practical capabilities, and explore its behavior and control in a wider range of environments, including on water surfaces.

\section{Conclusion}
\label{sec:Rollbot2-summary}

This paper presented Rollbot 2, a single-actuator spherical robot that combines a mechanically favorable rolling mode, onboard IMU/UWB sensing, and reinforcement-learning-based control. The platform supports commanded steering, onboard beacon seeking, and high-speed waypoint following through the same single motor interface. On the 4.8~m ``N'' trajectory, all 16 trials completed the path with a mean traversal time of 10.8~s and mean time-averaged deviation of 7~cm; the fastest trial was more than 70 times faster than the original Rollbot.

These results show that minimal actuation need not imply minimal behavior. By assigning complementary roles to morphology, sensing, estimation, and control, Rollbot 2 recovers useful agility and partial autonomy without adding actuators. More broadly, the comparison between the two Rollbot generations suggests a design strategy for underactuated robots: first identify a passive dynamic mode that makes the desired behavior accessible, then construct task-relative sensing and control around that mode.

\begin{singlespace}
\bibliographystyle{IEEEtran}
\bibliography{reference}
\end{singlespace}

\vfill

\end{document}